\documentclass{article}
\usepackage{graphicx}
\usepackage{spconf}
\usepackage{amsmath}
\usepackage{multirow}
\usepackage{booktabs}
\usepackage{amssymb}
\usepackage{amsfonts}
\usepackage{subcaption}
\usepackage{color}
\usepackage{algorithm}
\usepackage{algorithmic}
\usepackage{hyperref}
\usepackage{setspace}

\DeclareMathOperator*{\argmax}{argmax}

\title{U-TELL: UNSUPERVISED TASK EXPERT LIFELONG LEARNING}
\name{Indu Solomon$^{\star}$ 
\qquad Aye Phyu Phyu Aung$^{\dagger}$ 
\qquad Uttam Kumar$^{\star}$ \qquad Senthilnath Jayavelu$^{\dagger}$ 
\thanks{Indu Solomon and Uttam kumar  are grateful to International Institute of Information Technology Bangalore, India for the infrastructure support and acknowledges Mphasis Cognitive Computing Centre of Excellence for the financial assistance under Grant No. 7111.}
\thanks{Aye Phyu Phyu Aung and Senthilnath Jayavelu acknowledge funding from the Accelerated Materials Development for Manufacturing Program at A*STAR via the AME Programmatic Fund by the Agency for Science, Technology and Research under Grant No. A1898b0043.}
}
\vspace{-0.8cm}
\address{$^{\star}$International Institute of Information Technology Bangalore, India
\\
$^{\dagger}$Institute for Infocomm Research (I$^{2}$R), A*STAR, Singapore\\
}
\begin{document}
%
\maketitle
\begin{abstract}
Continual learning (CL) models are designed to learn new tasks arriving sequentially without re-training the network. However, real-world ML applications have very limited label information and these models suffer from catastrophic forgetting. To address these issues, we propose an unsupervised CL model with task experts called \textbf{U}nsupervised \textbf{T}ask \textbf{E}xpert \textbf{L}ifelong \textbf{L}earning (U-TELL) to continually learn the data arriving in a sequence addressing catastrophic forgetting.  
During training of U-TELL, we introduce a new expert on arrival of a new task. Our proposed architecture has task experts, a structured data generator and a task assigner. Each task expert is composed of 3 blocks; i) a variational autoencoder to capture the task distribution and perform data abstraction, ii) a $k$-means clustering module, and iii) a structure extractor to preserve latent task data signature. During testing, task assigner selects a suitable expert to perform clustering. U-TELL does not store or replay task samples, instead, we use generated structured samples to train the task assigner. We compared U-TELL with five SOTA unsupervised CL methods. U-TELL outperformed all baselines on seven benchmarks and one industry dataset for various CL scenarios with a training time over 6 times faster than the best performing baseline.
\end{abstract}

%
\begin{keywords}
unsupervised, continual learning, task, expert, structure
\end{keywords}
%
\vspace{-0.5cm}
\section{Introduction}
\vspace{-0.3cm}
\label{sec:intro}
Natural learning systems of living organisms can continually learn and adapt to the changes in the environment, but typical ML models \cite{parisi2019continual} suffer from catastrophic forgetting or inference, where newly acquired knowledge interferes with past learning \cite{parisi2019continual}. Most ML models solely depend on re-training to acquire new knowledge. Three scenarios of continual learning are: (i) task incremental learning (Task-IL) where the model incrementally learns a set of tasks with known task identity, e.g. learning a different musical instrument. (ii) class incremental learning (Class-IL) where the model learns new tasks with different classes, e.g., learning a new task incrementally with labels (car, truck), post learning the previous task with labels (cat, dog), (iii) domain incremental learning (Domain-IL) where the problem structure of the model is always the same but the input distribution varies \cite{van2022three}, e.g., identifying the same objects in different lighting conditions. In Domain-IL, every task has the same classes, and while testing, the task identity is not known \cite{de2021continual,van2022three}.


Supervised, unsupervised and self-supervised settings are commonly used with continual learning. Majority of research has gone into supervised continual learning. 
Data privacy and high labeling cost issues make supervised continual learning highly impractical. Unsupervised continual learning is more suitable for real-life scenarios with limited or no labeled data.
Among unsupervised continual learning models, KIERA \cite{pratama2021unsupervised} uses a network-growing architecture along with replay to counter catastrophic forgetting and takes longer time to train. UPL-STAM \cite{smith2021unsupervised} performs clustering at a heirarchy of receptive fields with two memories for the storage of cluster centroids. 
On the other hand, performance of self-supervised models like SCALE \cite{yu2023scale} and CaSSLe \cite{fini2022self} deteriorates in the unsupervised setting with unknown task boundaries.

In this paper, we propose \textbf{U}nsupervised \textbf{T}ask \textbf{E}xpert \textbf{L}ifelong \textbf{L}earning (U-TELL)  to address the issues with the existing methods.
U-TELL uses a structure-growing architecture, and hence, is free from the catastrophic forgetting issue \cite{parisi2019continual, aljundi2017expert}. U-TELL has a task assigner ($TA$) to assign the test samples to a suitable expert. Moreover, U-TELL does not store any task data samples for replay and trains faster compared to the chosen baselines, i.e., both computation and memory efficient. The main contributions are summarized as three-fold: 

\noindent 1. A memory efficient unsupervised CL model with task experts to learn the task data distribution and perform clustering in low dimensional latent space.

\noindent 2. Storage of low dimensional latent task distribution structure in place of task data samples and generation of samples to train the task assigner module.  

\noindent 3. Cross-entropy based task expert selection for Class-IL SMNIST, SCIFAR10, SSVHN, SCIFAR100, STinyImagenet and semiconductor industry Wafer defect map data, as well as cosine similarity based task expert selection for Domain-IL RMNIST and PMNIST data.    

Finally, we provide a comprehensive evaluation and ablation studies on the performance of our U-TELL architecture.
Experiment results show that U-TELL outperforms all the selected SOTA baselines and achieves both better time and memory efficiency against the baselines with both unsupervised and self-supervised settings for clustering task.

\vspace{-0.3cm}
\section{Related Works}
\noindent\textbf{Supervised Continual Learning}. Replay based \cite{rebuffi2017icarl, rolnick2019experience, isele2018selective, chaudhry2018efficient,chaudhry2019continual, lopez2017gradient, ebrahimi2020adversarial} 
and regularization based  \cite{li2017learning, kirkpatrick2017overcoming} methods suffer from catastrophic forgetting. 
Structure-based models are based on network architecture growth  \cite{mallya2018packnet, mallya2018piggyback, rusu2016progressive, aljundi2017expert, serra2018overcoming} and this strategy assures zero catastrophic forgetting at the cost of increased network parameters. 
Expert Gate \cite{aljundi2017expert} does not store samples and has an autoencoder-based gate to check for the most relevant model \cite{chen2018continual}. However, it requires multiple  autoencoder gates equal to the total number of tasks and thus, is unfavorable for unsupervised scenarios. 
In LwF \cite{li2017learning}, nodes are added to output layer on arrival of a new task. Soft labels of the old task model for the new data are used in the objective function \cite{chen2018continual}. Nevertheless, low task relatedness among task sequences can cause this shared model to perform worse in clustering jobs with more than two tasks \cite{aljundi2017expert}. 
\noindent\textbf{Self-supervised Continual Learning}. CL with Self-Supervised Learning (SSL) counter catastrophic forgetting through contrastive learning. 
In CaSSLe \cite{fini2022self} the self-supervised loss functions are converted to knowledge distillation for CL by adding an extra predictor network that maps the current state of the representations to their past state. 
Nevertheless, CaSSLe is applicable only in settings with known task boundaries and computation \& memory intensive \cite{fini2022self}. SCALE \cite{yu2023scale} trains in an unsupervised manner and is based on a pseudo-supervised contrastive loss, a self-supervised forgetting loss and a uniform subset selection-based replay buffer. 
However, the performance is dependent on the selected backbone SSL architectures and the stored samples in the memory buffer.

\noindent\textbf{Unsupervised Continual Learning}. UPL-STAM \cite{smith2021unsupervised} 
learning is based on a sequence of centroid learning, clustering, novelty detection, forgetting outliers, and storing important centroids in place of task exemplars. However, UPL-STAM uses two memories for storing the centroids \cite{smith2021unsupervised}. Yet, it is not a neural network model and the train and test images have to go through a pre-processing stage. Moreover, UPL-STAM works better with grayscale images. Unsupervised continual learning via pseudo labels \cite{he2021unsupervised} uses pseudo labels of $k$-means clustering instead of ground truth for training the model. But the algorithm demands the first task to be trained in a supervised manner and does not address the catastrophic forgetting issue. KIERA \cite{pratama2021unsupervised} is a self-evolving deep clustering network. KIERA has a self generated flexible network structure with a stacked autoencoder backbone and address the forgetting issue through experience replay.
However, the training time requirement in KIERA is unreasonably high and the performance drops heavily for real-world datasets like SCIFAR10, SSVHN and SWAFER.

\noindent\textbf{Motivation.
} 
Our proposed method is motivated by the modularity theory of human brain. 
Lesions in specific areas of brain can cause impairment of particular functionalities \cite{seger2010category}.
 If there is only a single network in our brain handling all functions, we will suffer catastrophic forgetting issue, where the discrimination between new knowledge and past experience is reduced as the network learns more. 
To prevent forgetting, a stimulus is sent to the concerned functional area whenever our brain is exposed to any new experience \cite{russo2014brain, seger2010category}. Likewise, we propose U-TELL: Unsupervised Task Expert Lifelong Learning model with respective experts for each incoming task to effectively counter the catastrophic forgetting issue.  

\label{sec:relatedworks}
\begin{figure}[!ht]
\centering
\fontsize{7}{9}\selectfont
\begin{subfigure}[b] {0.47\textwidth}
         \centering         \includegraphics[width=0.95\linewidth]{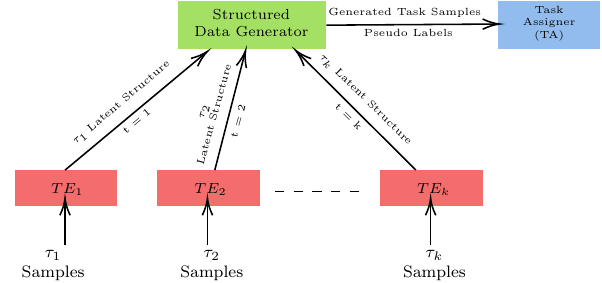}
         \captionsetup{font=footnotesize}
         \caption{}
         \label{fig:fullblockdia }
\end{subfigure}
\begin{subfigure}[b] {0.27\textwidth}
            \centering
\includegraphics[width=1\linewidth]{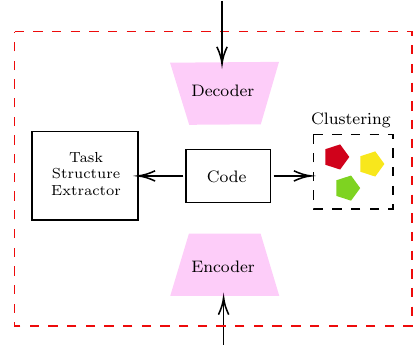} 
         \captionsetup{font=footnotesize}
         \caption{}
         \label{fig:expert }
         
\end{subfigure}
\begin{subfigure}[b] {0.32\textwidth}
            \centering
\includegraphics[width=1\linewidth]{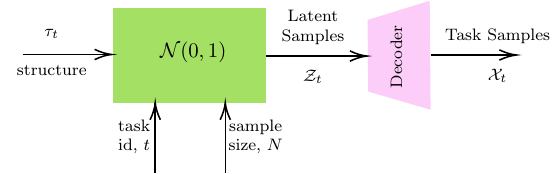}       
         \captionsetup{font=footnotesize}
         \caption{}
         \label{fig:structgen }
\end{subfigure}
\begin{subfigure}[b] {0.15\textwidth}
            \centering
\includegraphics[width=1\linewidth]{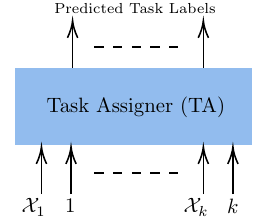}       
         \captionsetup{font=footnotesize}
         \caption{}
         \label{fig:tasngr }
         
\end{subfigure}
\captionsetup{font=footnotesize}
\caption{Structure of U-TELL architecture: (a) Block diagram; (b) Detailed diagram of each task expert; (c) Structured data generator in detail; (d) Input and output of the task assigner.}
\label{fig:fig1}
\end{figure}
\vspace{-0.7cm}
\section{U-TELL Architecture}
\label{sec:method}
\vspace{-0.3cm}
\subsection{Problem setup}
Let $t=1,2, \dots, k$ indicate the consecutive tasks. Given a sequence of tasks $\mathcal{T}$, each task $\tau_{t}$ has $n_{t}$ training sample images i.e, $X_{t} = x_j, {j=1,\ldots, n_{t}}$ and $x_j \in \mathbb{R}^{n \times m \times l}$. 
Latent space samples are denoted by $Z_{t}$, where $Z_{t} \in \mathbb{R}^d$ and $d \ll n \times m \times l$. The lowercase notations ($x_j$, $z_j$, etc.) denote the individual task data samples. 
Generated structured latent samples are denoted by $\mathcal{Z}_t$ and data space structured samples are denoted by $\mathcal{X}_t$. Generated pseudo labels are represented by $C_{t}$ and $Y_{t}$ denote true labels. We analyze three common continual learning frameworks, viz., (a): task incremental learning (Task-IL) (b): domain incremental learning (Domain-IL), and (c): class incremental learning (Class-IL) \cite{cha2021co2l}. 
We train U-TELL on a sequence of tasks, $\mathcal{T}$. In Task-IL and Class-IL setup, no two tasks will have the same class labels i.e., their sample spaces are disjoint, 
$t\neq {t^{\prime}} \Rightarrow Y_t\cap Y_{t^{\prime}} = \emptyset.$
\normalsize
In Domain-IL the class labels for all the tasks are the same, 
$Y_{t} = Y_{t^{\prime}}.$
\normalsize
Task expert and task assigner are denoted by $TE$ and $TA$ respectively.  
\vspace{-0.3cm}
\subsection{Proposed Architecture}




We present the architecture of U-TELL in Figure~\ref{fig:fig1}. Figure \ref{fig:fullblockdia } shows the block diagram of the training process with three major blocks namely the task expert ($TE$), the structured data generator ($SDG$), and the task assigner ($TA$). The task expert is a flexible module, added on arrival of a new task. The structured data generator block generates synthetic task samples from the stored task structure signatures. The task assigner is meant for guiding the test samples to an appropriate task expert. The overall skeleton of the U-TELL algorithm is shown in Algorithm \ref{alg:uel_alg}.
\begin{algorithm}[tb]
\small
\caption{Unsupervised Task Expert Lifelong Learning (U-TELL)}
\label{alg:uel_alg}
\textbf{Input}: Task data- $\{X_{t}\}$, $t=1,\ldots,k$\\
\textbf{Output}: Trained network of task experts- $\{TE_t\}$, $t = 1,\ldots,k$ and 
task assigner- $TA$
\begin{algorithmic} [1] 
\STATE Let $X_{t}$, $t=1, \dots, k$ be the task data arriving continually, 
\WHILE{$t$}
\STATE Train a new expert $TE_t$ with $X_{t}$ and get latent data $Z_t$
\STATE Perform clustering on latent data $Z_t$
\STATE Compute covariance structure $Q_t$ and eigenspectrum $D_t$ and $V_t$ for task $t$
\ENDWHILE
\STATE Use $SDG$ module to generate structured samples and pseudo labels$\{(\mathcal{Z}_t,C_t)\}$, $t=1,\ldots,k$ and transform to $\{\mathcal{X}_t\}$, $t=1,\ldots,k$
\STATE Train task assigner $TA$ with $\{(\mathcal{X}_t,C_t)\}$, $t=1,\ldots,k$
\end{algorithmic}
\end{algorithm}

\noindent\textbf{Task Expert.}
Task Experts ($TE$) are the core of U-TELL architecture. We present the detailed architecture of the task expert in Figure \ref{fig:expert }. Three main blocks of task expert are variational autoencoder (VAE), $k$-means clustering block, and task structure extractor block. The encoder-decoder networks of VAE help in learning the task data distribution and bring down the dimensionality of data.
$X_t$ and $Z_t$ represent the input and latent data respectively.
\normalsize

The $k$-means clustering operates on the latent data $Z_t$ and forms meaningful clusters. The task structure extractor module estimates the structure of task data distribution and preserves it in memory. The required number of $TE$s is unknown at the start of the training and is dynamically allotted, as new tasks arrive on the fly. Knowledge sharing among $TE$s is ensured by weight initialization of the current block with the previous block's learned weights. This process guarantees faster convergence as the new $TE$s do not learn from scratch. 
Let $Z_t$ and $\mu_t$ be the sample array and sample mean of $t^{th}$ latent data distribution. The covariance structure of latent distribution $Q_t$ is given by, 
\vspace{-0.1cm}
\small
\begin{equation}\label{cov_eq}
Q_{t}  = \frac{1}{N} \sum( Z_{t}-\mu_{t}) (Z_{t}-\mu_{t})^T.
\end{equation}
\normalsize
Eigenspectrum of the covariance structure $Q_t$ of $t^{th}$ task data is computed and stored in the memory for further processing in $SDG$ block. 

\noindent\textbf{Structured Data Generator.}
The Structured Data Generator ($SDG$) module contains two computational blocks viz., the data generator and the trained decoder. Figure \ref{fig:structgen } shows the block diagram of $SDG$. The inputs to the $SDG$ block are the latent task data distribution signature ($D_t$, $V_t$), the task id $t$, and the required sample size $N$. 
The latent task data distribution signature and decoder network are loaded from memory based on the input task $t$. The required number of samples are sampled from a standard normal distribution $\mathcal{N}(0,I)$ and mapped to the required covariance structure with the help of task signature. 
\vspace{-0.1cm}
\small
\begin{equation}\label{trans_eqn}
I = (V_t D_t^{-1/2})^T Q_{t} (V_t D_t^{-1/2}) = F_t^T Q_{t} F_t,
\end{equation}
where $F_t = V_t D_t^{-1/2}$ has a unit covariance structure.
\vspace{-0.1cm}
\small
\begin{equation}\label{sampling}
\mathcal{Z}_t = F_t^{-1}\mathcal{P}(\mathcal{N}(0,I)),
\end{equation}
\normalsize
where $\mathcal{P}(\cdot)$ represents the random sampling process. 
The generated latent space samples $\mathcal{Z}_t$ are passed through the chosen decoder to obtain the required task data samples.
\vspace{-0.1cm}
\small
\begin{equation}\label{data_samples}
\mathcal{X}_t = Decoder_t(\mathcal{Z}_t)
\end{equation}
\normalsize
\noindent\textbf{Task Assigner.}
 The main role of the Task Assigner ($TA$) module is to predict a suitable expert for the test data. Figure \ref{fig:tasngr } illustrates the $TA$ module.
Inputs to the $TA$  module are structured samples and pseudo labels. Pseudo labels are our algorithm generated labels used to identify the tasks and not related to the true labels of training data samples. We propose two variations of $TA$ namely,  cross entropy-based - $TA_{CE}$ and cosine similarity-based - $TA_{cos}$. In Class-IL, a sequence of tasks of different class labels are classified by $TA_{CE}$ and in Domain-IL, a sequence of tasks of same class labels are categorized by $TA_{cos}$. In Domain-IL different task inputs come from the drifted versions of original task distribution,
$Y_t = Y_{t+1}$ and $P(t) = P(t+1)$, where $P(\cdot)$ is task input data distribution. In Class-IL the input task data distributions are different $Y_t \subset Y_{\mathcal{T}}$ and $P(t) \ne P(t+1)$.


\noindent\textbf{Training Procedure.} U-TELL training involves two steps: (i) training of Task Experts ($TE$)s, (ii) generation of structured samples and training of Task Assigner ($TA$). As a new task arrives, a new $TE$ is initialised and trained to handle the new task. On training completion of each $TE$, the encoder-decoder, $k$-means model, and task distribution signature are stored in the memory. 
The second part of the training process starts with the generation of synthetic data. The latent structure signature is chosen based on the task and the required number of samples are sampled from a normal distribution. These unit variance samples are mapped to the required covariance structure with the help of stored task data distribution signature. Latent space samples are passed through the selected decoder to obtain data space samples. Pseudo labels for $TA$ training are generated. Pseudo labels can be retained the same as task $t$ or other variable names can be assigned and accordingly, $TE$s have to be identified. Generated samples and pseudo labels are passed on to $TA$.

U-TELL does not apply restrictions either on the number of tasks or on the number of classes per task. 
As U-TELL architecture is modular, it introduces new $TE$s in the same sequential manner upon the reception of tasks, $t_{1}, ..., t_{k}$. After receiving $k^{th}$ task and training $TE_k$, the algorithm continues with the generation of structured samples, $\mathcal{Z}_t$, $\mathcal{X}_t$, followed by training $TA$ to complete the training process. 
\vspace{-0.3cm}
\subsection{Training Objective}
Training objective of U-TELL involves the minimization of 3 losses. The overall training objective of U-TELL is given by,
\vspace{-0.1cm}
\small
\begin{equation}\label{total_loss}
 \mathcal{L} = \lambda_1 \sum_t \mathcal{L}_{TE}+\lambda_2  \sum_t \mathcal{L}_{struct}+\lambda_3 \mathcal{L}_{TA},
 \end{equation}
 \normalsize
 where $\lambda_{1,2,3}$ are tunable parameters. The first term $\mathcal{L}_{TE}$ of 
 Eq. (\ref{total_loss}) is the $TE$'s training objective for each task $t$ and is given by, 
 \small
\begin{equation}\label{te_loss}
\mathcal{L}_{TE} = \mathcal{L}_{\beta_{vae}}+\mathcal{L}_{clust},  \\
 \end{equation}
\normalsize
where $\mathcal{L}_{\beta_{vae}}$ is the encoder-decoder loss of disentangled VAE, also called beta-VAE \cite{higgins2016beta}.
beta-VAE is a modified version of vanilla VAE and can provide better disentanglement of distribution in the latent space \cite{gong2018learning}.
The beta-VAE training objective is given by,
\vspace{-0.3cm}
\small
\begin{multline}\label{beta_vae}
 \mathcal{L}_{\beta_{vae}}(\theta,\phi) = \sum_{j\in X_t}\mathbb{-E}_{z\sim Q_\theta(z|x_{j})}\big[ \log P_\phi(x_{j}|z)\big] \\ + \beta \cdot KL \big( Q_\theta(z|x_{j}) \Vert P(z) \big),  
\end{multline}
\normalsize
where $\theta$ and $\phi$ refer to the parameters of encoder and decoder blocks respectively, $\beta$ is the disentanglement term and is a tunable hyperparameter. $\beta > 1.0$ puts a stronger constraint on latent data and thus, the model is compelled to learn the best representation of the data \cite{gong2018learning}. The first term of $\mathcal{L}_{\beta_{vae}}$ from Eq. (\ref{beta_vae}) is a reconstruction loss, where $\mathbb{E}[\cdot]$ is the expected value of decoder output $x_j$ and the second term is the KL divergence between the encoder distribution and the standard normal distribution. The goal of the encoder is to learn the probability distribution $Q_{\theta}(z|x_{j})$, given the input data $X_{t}$. The goal of the decoder is to learn the probability distribution $P_{\phi}(x_{j}|z)$. $P(z)$ represents the standard normal distribution. The KL divergence term in the loss function helps in pushing the learned encoder distribution $Q_{\theta}(z|x_{j})$ as close as possible to the standard normal distribution. The term $\mathcal{L}_{clust}$ in Eq. (\ref{te_loss}) is the clustering loss. The clustering loss is given by,
\vspace{-0.1cm}
\small
\begin{equation}\label{cluster_loss}
 \mathcal{L}_{clust} = 
 \sum_{n=1}^{N}\sum_{r=1}^{R}{{w_{nr}}\left \Vert {Z_{n}-\mu_{r}}\right \Vert ^2},   \\
 \end{equation}
\normalsize
where $R$ represents the total number of clusters formed for each $TE$, $N$ represents  the task sample count for each $TE$, $w_{nr}$ is the cluster membership variable and it takes a value of 0 or 1. $Z$ and $\mu$ represent the latent samples and cluster center respectively.

The second term $\mathcal{L}_{struct}$ of Eq. (\ref{total_loss}) is the loss minimized in the generation of structured task samples and is given by,
\small
\begin{equation}\label{struct_loss}
 \mathcal{L}_{struct} = 
\sum_{r}{\mathfrak{D}}(Z_{{r}},\mathcal{Z})-\sum_{r^{\prime}}\mathfrak{D}(Z_{{r^{\prime}}},\mathcal{Z}),  \\
 \end{equation}
\normalsize
where $Z_{{r}}$ is the latent task data cluster, $\mathcal{Z}$ is the generated latent sample and $\mathfrak{D} (., .)$ is a distance function. $\mathcal{L}_{struct}$ ensures the generated latent task samples are closer to the original latent data cluster $r$ and far from other irrelevant clusters $r^{\prime}$. 

The third loss in Eq. (\ref{total_loss}) is the $TA$ loss, $\mathcal{L}_{TA}$. $TA$ loss for Class-IL is a cross-entropy loss function between the predicted task label and the pseudo task label.
\vspace{-0.1cm}
\small
\begin{equation}\label{ta_loss1}
 \mathcal{L}_{TA} = 
-\sum_{i}c_i\log(e_i) ,  \\
 \end{equation}
 \normalsize
where $c_i$ is the pseudo label and $e_i$ is the predicted task expert label. $TE$ selection for Domain-IL is performed by $TE$ suitability factor ($\mathfrak{T}_{t}$) computation. $\mathfrak{T}_{t}$  is the sum of cosine similarity index between the generated structured task data samples and the test samples and is given by,
\vspace{-0.1cm}
\small
\begin{equation}\label{ta_loss2}
 \mathfrak{T}_{t} = 
\sum_{l}\frac{{\mathcal{X}_t \cdot X_l}}{\left \Vert \mathcal{X}_t\right \Vert \cdot \left \Vert X_l\right \Vert}, \forall{t},\\
 \end{equation}
\normalsize
 where $X_l$ denotes the test samples array. Cosine similarity varies between 0 and 1. Data samples closer to each other have cosine similarity values closer to 1 and samples farther from each other have cosine similarity closer to 0. The selection for an appropriate $TE$ is given by,
 \vspace{-0.1cm}
 \small
\begin{equation}\label{te_select}
TA_{cos} = \argmax_{t}\big( \mathfrak{T}_{t} \big),
\vspace{-0.3cm}
\end{equation}
\normalsize
where $\argmax(.)$ function gives the $TE$ index, $t$, corresponding to the max value of $\mathfrak{T}_{t}$. 

\vspace{-0.3cm}
\section{Experimental Results}
\begin{table*}[h!]
\centering
\fontsize{7}{9}\selectfont
\setlength{\tabcolsep}{3pt}
\captionsetup{font=footnotesize, justification= centering}
\caption{Performance comparison of U-TELL with other baselines for SMNIST, SCIFAR10, SSVHN, \\ SCIFAR100, STinyImagent, RMNIST and PMNIST datasets}
\begin{tabular}{ccccccccc}
\hline
\multirow{2}{*}{\textbf{Technique}} & \multicolumn{6}{c}{\textbf{Dataset}} \\ \cmidrule(l){2-9}
& \multicolumn{1}{c}{SMNIST} & \multicolumn{1}{c}{SCIFAR10}  
&\multicolumn{1}{c}{SSVHN} &
\multicolumn{1}{c}{SCIFAR100}&
\multicolumn{1}{c}{STinyImageNet}&
\multicolumn{1}{c}{SWAFER} &
\multicolumn{1}{c}{RMNIST} & \multicolumn{1}{c}{PMNIST}\\
\hline
LwF  &61.19$\pm$1.35 &25.30$\pm$1.10 &16.06$\pm$0.44 &-- &--&51.35$\pm$2.79 &55.09$\pm$1.94  &47.51$\pm$3.48\\
KIERA  &84.81$\pm$1.44 &24.51$\pm$1.01 &15.03$\pm$0.33 &-- &-- &37.40$\pm$3.72 &80.04$\pm$0.28 &66.86$\pm$2.88\\
UPL-STAM  &86.44$\pm$1.22 &28.02$\pm$1.99  & 56.34$\pm$2.11 &13.22$\pm$0.32 &21.68$\pm$1.30 &-- &-- &--  \\
CaSSLe  &45.01$\pm$2.41 &16.91$\pm$1.00  &14.93$\pm$1.06 &10.97$\pm$0.84 &20.66$\pm$1.35 &-- &-- &--  \\
SCALE  &52.84$\pm$4.99 &21.72$\pm$1.01  &14.91$\pm$1.19 &13.68$\pm$0.78 &21.66$\pm$0.68 &-- &-- &--  \\
\textbf{U-TELL (ours)} &\textbf{87.68$\pm$0.13} &\textbf{29.65$\pm$0.67}  &\textbf{58.43$\pm$0.79}
&\textbf{16.79$\pm$0.38}
&\textbf{27.78$\pm$1.13}
&\textbf{60.18$\pm$2.78} &\textbf{85.91$\pm$0.38} &\textbf{86.09$\pm$0.72}\\
\hline
\end{tabular}
\label{tab:tableavgacc} 
\end{table*}

\begin{figure}[t]
\centering
\fontsize{7}{9}\selectfont
\includegraphics[width=0.49\textwidth]{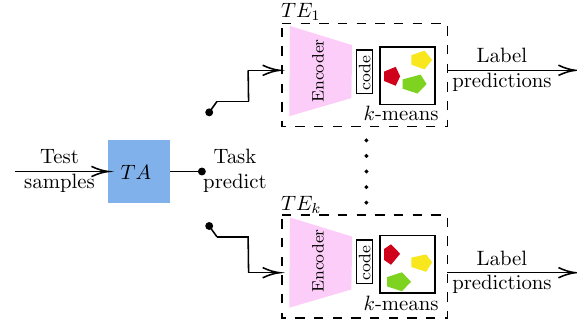} 
\captionsetup{font=footnotesize}
\caption{Block diagram of the testing phase of the proposed U-TELL.}
\label{Fig2_testingblockdia}
\end{figure}
\vspace{-0.2cm}

\noindent\textbf{Testing Procedure.}
Test samples are passed through the task assigner ($TA$) module to identify a suitable task expert ($TE$) for the test data. The encoder block of the selected $TE$ is fed with the test samples to obtain the latent data and the latent test samples are further fed to the $k$-means block of $TE$ to obtain the cluster predictions. These clusters are mapped to class labels to obtain the final model predictions. 
Figure \ref{Fig2_testingblockdia} describes the testing phase of the proposed U-TELL. 

\noindent \textbf{Learning Settings.}
We evaluate our U-TELL in an unsupervised setting for Class-IL and Domain-IL frameworks. We assume unknown task boundaries and use our $TA$ model to predict the suitable $TE$. 


\noindent\textbf{Datasets.} We evaluate U-TELL for 8 prominent CL datasets namely split MNIST \cite{lecun1998mnist} (SMNIST), split CIFAR10 (SCIFAR10) \cite{krizhevsky2009cifar, krizhevsky2014cifar}, split SVHN (SSVHN) \cite{netzer2011reading}, split CIFAR100 (SCIFAR100) \cite{krizhevsky2014cifar}, split TinyImagenet (STinyImagenet) \cite{deng2009imagenet}, split WAFER  (SWAFER) defect \cite{wang2020deformable}, rotated MNIST (RMNIST) \cite{ghifary2015domain}, and permuted MNIST (PMNIST) \cite{goodfellow2013empirical} \footnote{Data processing steps, implementation settings and source code is shared at \url{https://github.com/indusolo/Unsup_LL.git}.}. 

\noindent\textbf{Baselines.}
We select UPL-STAM \cite{smith2021unsupervised}, KIERA \cite{pratama2021unsupervised}, SCALE \cite{yu2023scale}, CaSSLe \cite{fini2022self}, and LwF \cite{li2017learning} 
as comparison baselines for our experimental study. We remove task label information to adapt CaSSLe \cite{fini2022self} to the unsupervised CL setting. We ensure the unsupervised setting for LwF using the same backbone setting of VAE and clustering with unlabeled data.
We have ensured fairness in our study by running the baselines' publicly available codes in the same computational environment as U-TELL. We chose $ACC$ as evaluation metric, $ ACC = \frac{1}{k}\sum_{t=1}^k A_{t}$, 
\normalsize
where $k$ is the total number of tasks and $A_t$ is the test classification accuracy for task $t$.

\begin{figure*}[t]
\centering
\fontsize{7}{9}\selectfont
\begin{subfigure}[b]{.32\linewidth}
\fontsize{7}{9}\selectfont
\includegraphics[width=\textwidth]{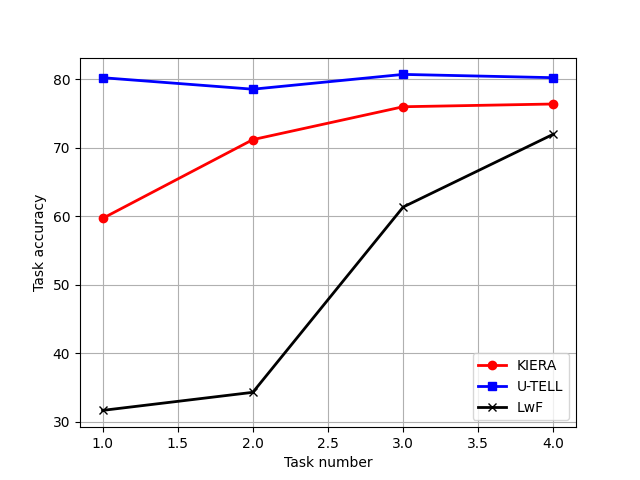} 
\captionsetup{font=footnotesize}
\caption{} 
\label{fig:pm_tsk_acc}
\end{subfigure}
\hspace{0.4cm}
\begin{subfigure}[b]{.32\linewidth}
\fontsize{7}{9}\selectfont
    \centering
\includegraphics[width=\textwidth]{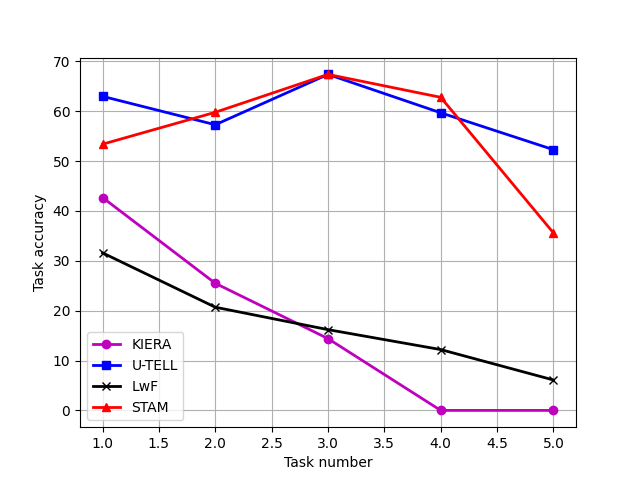} 
\captionsetup{font=footnotesize}
\caption{}
\label{fig:ssvhn_tsk_acc}
\end{subfigure}
\begin{subfigure}[b]{.32\linewidth}
    \centering
\fontsize{7}{9}\selectfont
\includegraphics[width=\textwidth]{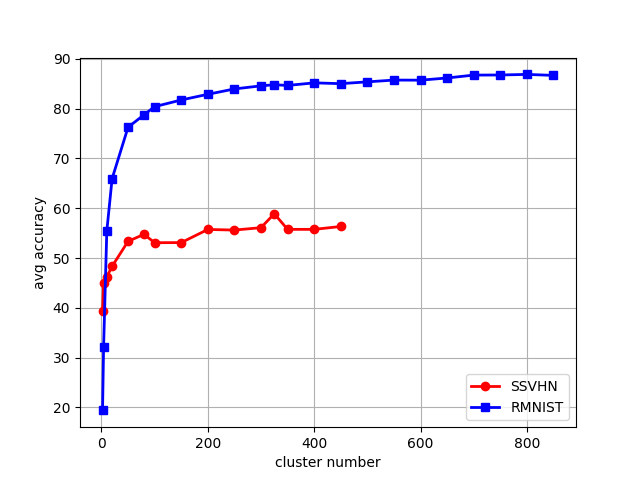} 
\captionsetup{font=footnotesize}
\caption{}
\label{fig:abln}
\end{subfigure}
\vspace{-0.3cm}
\captionsetup{font=footnotesize}
\caption{(a) Comparison of individual task performance of U-TELL with the baselines for PMNIST dataset; (b) Comparison of individual task performance of U-TELL with the baselines for SSVHN dataset; (c) U-TELL ablation study conducted by varying the cluster numbers for SSVHN and RMNIST dataset.} 
\end{figure*}
\noindent\textbf{Comparison.}
In Table \ref{tab:tableavgacc}, we present the comparison results of U-TELL and the chosen baselines. We can see that in both Class-IL and Domain-IL settings, our proposed U-TELL has outperformed all the baselines by large margins. In Domain-IL experiments, RMNIST and PMNIST show an improvement of 7.33$\%$ and 28.11$\%$ respectively compared to the best performing baseline.
Class-IL experiments on SMNIST, SCIFAR10, SSVHN, SCIFAR100, STinyImagenet, SWAFER datasets show a rise in average accuracy of 1.43$\%$, 5.82$\%$, 3.71$\%$, 22.73$\%$, 28.14$\%$  and 12.23$\%$ respectively.


Figure \ref{fig:pm_tsk_acc} presents the performance comparison of U-TELL with baselines for the PMNIST dataset. In this case, we can see that KIERA and LwF suffer from catastrophic forgetting for older tasks and perform poorly on them. 
In the case of U-TELL, $TE$s are able to give optimal performance for all four tasks without catastrophic forgetting. Figure \ref{fig:ssvhn_tsk_acc} displays the performance comparison for the SSVHN dataset. The image quality issues of SSVHN make it tougher to classify compared to other digit datasets such as MNIST. KIERA and LwF perform poorly on SSVHN. U-TELL gives the best performance on the SSVHN dataset and among the chosen baselines, UPL-STAM's performance is closer to U-TELL. Figures \ref{fig:pm_tsk_acc} and \ref{fig:ssvhn_tsk_acc} verify that the U-TELL is not affected by catastrophic forgetting issues and is able to perform well on all the sequential tasks. 
\vspace{-0.3cm}
\begin{table}[h!]
\centering
\fontsize{7}{9}\selectfont
\setlength{\tabcolsep}{3pt}
\captionsetup{font=footnotesize}
\caption{Training efficiency comparison. Percentage information in magenta indicates the efficiency compared to U-TELL.}
\begin{tabular}{cccc}
\toprule
\multirow{2}{*}{\textbf{Technique}}  & \multicolumn{1}{c}{Training time}  
&\multicolumn{1}{c}{Training time.}  \\
& \multicolumn{1}{c}{(SMNIST)}  
&\multicolumn{1}{c}{(RMNIST)}  \\
\midrule
LwF &875 \textcolor{magenta}{(+150\%)}  &870 \textcolor{magenta}{(+202\%)} \\
KIERA &20725 \textcolor{magenta}{(+3573\%)} &10600 \textcolor{magenta}{(+2465\%)} \\
UPL-STAM &3689 \textcolor{magenta}{(+636\%)} &-- \\
CaSSLe &835 \textcolor{magenta}{(+144\%)}  &--\\
SCALE &867 \textcolor{magenta}{(+150\%)}  &--\\
U-TELL (Ours) &\textbf{580} &\textbf{430} \\
\bottomrule
\end{tabular}
\label{tab:train_efficiency} 
\end{table}

\noindent\textbf{Training Efficiency.}
Table \ref{tab:train_efficiency} presents a training time efficiency comparison of U-TELL with the selected baselines for SMNIST and RMNIST datasets. We observe that our proposed U-TELL can handle multiple $TE$s and is able to perform the best with the lowest training time requirement.
U-TELL training time requirement for SMNIST and RMNIST are 536$\%$  and 2365$\%$ lesser than the best performing baseline. The lower training time requirement of U-TELL is advantageous to have a scalable architecture, which can be extended for a large number of tasks with a reasonable training time and uniform performance.

\noindent\textbf{Memory Efficiency.} Table \ref{tab:mem_eff} presents the memory requirements in Megabytes (MB) for SCALE, UPL-STAM and KIERA against our U-TELL on SMNIST and SSVHN datasets.
The proposed model stores the compressed latent space signatures of the task distributions. Hence, the memory requirement is dependent on the latent space dimension. The memory requirement is given by,
\vspace{-0.1cm}
\small
\begin{equation}\label{uexlmem}
 M = 
2(dkn_k)+d^2kn_k,\\
 \end{equation}
\normalsize
\vspace{-0.1cm}
\noindent where $d$ refers to latent space dimensionality, $k$ is the total number of tasks and $n_k$ is the number of classes per task. Given a memory size of 10MB and $d = 128$, $n_k = 2$, theoretically our model can handle upto 300 sequential tasks while baselines have exponential growth in memory requirement. 

\begin{table}[!h]
\centering
\fontsize{7}{9}\selectfont
\setlength{\tabcolsep}{3pt}
\captionsetup{font=footnotesize}
\caption{Memory efficiency comparison in MB}
\begin{tabular}{cccccc}
\hline
\multirow{2}{*}{} & \multicolumn{4}{c}{\textbf{Method}}  \\ \cmidrule{2-6} & \multicolumn{1}{c}{SCALE} & \multicolumn{1}{c}{UPL-STAM}  
&\multicolumn{1}{c}{KIERA} & \multicolumn{1}{c}{U-TELL} & \\
\hline
SMNIST &4.1 \textcolor{magenta}{(+2312$\%)$} &4.72 \textcolor{magenta}{(+2676$\%$)}    &3.2 \textcolor{magenta}{(+1782$\%$)} &\textbf{0.17}  \\
SSVHN &15.73 \textcolor{magenta}{(+2248$\%$)} &12.01 \textcolor{magenta}{(+1693$\%$)} &28.26 \textcolor{magenta}{(+4118$\%$)} &\textbf{0.67} \\
\hline
\end{tabular}
\label{tab:mem_eff} 
\end{table}

\noindent\textbf{Ablation Study.} We conduct an ablation study to understand the effect of the number of clusters on U-TELL performance. Figure \ref{fig:abln} shows the performance improvement with increasing number of clusters for SSVHN and RMNIST. However, after reaching around 200 clusters the accuracy improvement is not significant. Hence, the number of clusters can be selected based on the application and availability of computational resources. The performance drops after reaching the optimum for SSVHN and shows a small drop in RMNIST. Table \ref{tab:ta_abln} demonstrates the ablation study of different $TA$ blocks on various datasets. In case of Domain-IL, the network learns the same problem in different situations and this appears as a drift in the  task data distribution. This domain shift is captured well with cosine similarity between the generated data and the test data. In Class-IL input task data distributions of subsequent tasks are independent with mutually exclusive classes and the network handles different problems in different situations for each task. Hence, in Class-IL the cosine similarity between dissimilar data does not consistently select the suitable $TE$. 

\begin{table}[!h]
\centering
\fontsize{7}{9}\selectfont
\setlength{\tabcolsep}{3pt}
\captionsetup{font=footnotesize}
\caption{Ablation study on $TA$ effectiveness}
\begin{tabular}{cccccc}
\hline
\multirow{2}{*}{\textbf{TA}} & \multicolumn{5}{c}{\textbf{Dataset}}  \\ \cmidrule{2-6} & \multicolumn{1}{c}{SMNIST} & \multicolumn{1}{c}{SCIFAR10}  
&\multicolumn{1}{c}{SSVHN} & \multicolumn{1}{c}{RMNIST} & \multicolumn{1}{c}{PMNIST}\\
\hline
$TA_{CE}$ &\textbf{87.68$\pm$0.13} &\textbf{29.65$\pm$0.67}  &\textbf{58.43$\pm$0.79} &58.09$\pm$1.09 &77.01$\pm$0.48\\
$TA_{cos}$ &58.00$\pm$10.94 &24.03$\pm$1.54 &13.84$\pm$1.00  &\textbf{85.91$\pm$0.38} &\textbf{86.09$\pm$0.72}\\
\hline
\end{tabular}
\label{tab:ta_abln} 
\end{table}

\noindent\textbf{Discussion.}
Our model adopts dynamic architecture and hence, can counter catastrophic forgetting \cite{parisi2019continual, aljundi2017expert}. The main challenge here is to identify the matching task head for the test samples. Our U-TELL negates catastrophic forgetting through the $TE$s and the best $TE$ for the test samples is identified with the help of $TA$. Our model trains fast, do not store task data samples and has very low memory requirement. U-TELL outperformed all the selected baselines for all real-world and industry datasets.
U-TELL surpassed the baselines in ten sequential tasks of CIFAR100 data. U-TELL architecture is scalable and can be extended to large sequences of tasks. Most of the chosen baselines' performance drops when there is an increased task flow. On the other hand, the brain theory-inspired modular architecture has helped U-TELL to perform uniformly well for all the tasks. The performance of SCALE is highly dependent on hyperparameter tuning, online memory update and sample selection. The assumption of task boundary knowledge in CaSSLe is highly impractical and the performance deteriorates heavily in the absence of task identity. UPL-STAM has a dependency on task data preprocessing and performs better on grayscale images. KIERA's training time 
is unrealistically high with over 3000 clusters.
\vspace{-0.3cm}
\section{Conclusion}
\vspace{-0.3cm}
In this paper, we propose Unsupervised Task Expert Lifelong Learning (U-TELL), a structure growing architecture, that has task experts and task assigner modules. Task experts are added on the arrival of a new task and the task assigner serves in the selection of a suitable $TE$ for the test samples. U-TELL is scalable and can be deployed for a large number of tasks. It is also free of catastrophic inference and can handle all three continual learning scenarios while not storing any task data in memory. We compared U-TELL with UPL-STAM, KEIRA, SCALE, CaSSLe, and LwF on benchmark continual learning datasets and one industry application - Wafer defect map dataset. We used unsupervised continual learning setting with unknown task boundaries for all our experiments. Our U-TELL has outperformed all selected baselines in terms of average accuracy and training time.

\vfill\pagebreak
\setstretch{0.2}
\bibliographystyle{IEEEbib}
\bibliography{unsup_ll}
\end{document}